\documentclass{article}
\usepackage{ijcai24}

\usepackage{times}
\usepackage{soul}
\usepackage{url}
\usepackage[hidelinks]{hyperref}
\usepackage[utf8]{inputenc}
\usepackage[small]{caption}
\usepackage{graphicx}
\usepackage{amsmath}
\usepackage{amsthm}
\usepackage{booktabs}
\usepackage{algorithm}
\usepackage{algorithmicx}
\usepackage[noend]{algpseudocode}
\usepackage[switch]{lineno}
\usepackage{subcaption}
\usepackage{array}
\usepackage{setspace}
\usepackage[usenames]{color} 
\definecolor{ToDoColor}{rgb}{0.1,0.2,1}

\usepackage{comment}

\title{Expected Work Search: Combining Win Rate and Proof Size Estimation}
\author{
Owen Randall
\and
Martin Müller\and
Ting-Han Wei\And
Ryan Hayward\\
\affiliations
University of Alberta\\
\emails
\{davidowe, mmueller, tinghan, hayward\}@ualberta.ca
}
\date{}

\begin{document}

\maketitle

\begin{abstract}

We propose Expected Work Search (EWS), a new game solving algorithm. 
EWS combines win rate estimation, as used in Monte Carlo Tree Search, with proof size estimation, as used in Proof Number Search.
The search efficiency of EWS stems from minimizing a novel notion of Expected Work, which predicts the expected computation required to solve a position.
EWS outperforms traditional solving algorithms on the games of Go and Hex. 
For Go, we present the first solution to the empty 5$\times$5 board with the commonly used positional superko ruleset.
For Hex, our algorithm solves the empty 8$\times$8 board in under 4 minutes. 
Experiments show that EWS succeeds both with and without extensive domain-specific knowledge.
\end{abstract}

\section{Introduction}

Efficient search algorithms are critical for finding solutions in large problem spaces. 
For example, Monte Carlo Tree Search was a key component in finding faster matrix multiplication algorithms \cite{alphatensor}. 
Search has also been shown to improve reasoning and problem solving capabilities of large language models \cite{treesofthought}.

Expected Work Search (EWS) is a novel search algorithm for two player zero sum games with perfect information. EWS utilizes estimates of both win rate and proof size in order to predict the expected amount of work that it will take to solve a given position\footnote{We call the complete game state a \textit{position}, including information such as the board configuration of a Go game and the move history where needed. Furthermore, we identify nodes in the search tree with the positions they represent.}. 
The Expected Work (EW) heuristic is used to direct the search towards smaller, less expensive solutions, and minimize the time wasted searching more complex wins, or losing variations.
We evaluate EWS by solving small-board positions from the two player perfect information games of Go and Hex.
This involves proving which player wins with optimal play, and creates a complete winning strategy which has a response for every possible opponent move.

EWS combines and extends elements of two popular algorithms, Proof Number Search (PNS) \cite{pns} and Monte Carlo Tree Search (MCTS) \cite{mcts}.
These algorithms use dynamically updated heuristics to efficiently find solutions, which can give them an edge over more traditional minimax-based programs such as $\alpha\beta$ solvers \cite{alphabeta}.
MCTS uses its win rate estimation to deeply search lines of play following the best moves found so far.  PNS estimates proof sizes, and seeks to minimizes proof or disproof numbers in order to quickly find a solution.
Both algorithms have strengths and weaknesses. 
MCTS makes no effort to find an easy solution in a short amount of time, while PNS can spend much time on searching deep forcing lines that do not work.

Our results show that EWS by combining win rate and proof size estimation into a single expected work heuristic, outperforms existing search algorithms in both Go and Hex.
EWS is able to solve the empty 8$\times$8 Hex board in under 4 minutes on modest hardware (a first player win), and is the first program to solve the empty 5$\times$5 Go board using positional superko rules (a first player full-board win).

\section{Background}

$\alpha\beta$ pruning \cite{alphabeta} enhances the basic minimax algorithm by using upper and lower score bounds $\alpha$ and $\beta$ to prune irrelevant lines of play.
Iterative deepening $\alpha\beta$ is an enhancement to the algorithm which allows it to find wins in the minimum number of moves \cite{iterative_deepening}.
Each iteration executes a depth-limited $\alpha\beta$ search, and non-terminal positions at the depth limit are evaluated by a heuristic.
The best moves found in the previous iteration are prioritized so that strong lines of play are searched first.
With an accurate heuristic, this results in more $\alpha\beta$ pruning, saving computation.

Proof Number Search \cite{pns} uses proof and disproof numbers as optimistic heuristic estimates of the remaining cost to complete a proof or disproof. The algorithm constructs an in-memory proof/disproof tree by iteratively expanding most promising nodes, and updating proof/disproof numbers in the tree. PNS searches for solutions in a best-first manner, in contrast with iterative deepening $\alpha\beta$. PNS can quickly find narrow but deep solutions containing long forced lines of play, wit few branches \cite{pns_survey}. However, this depth-seeking behavior of PNS can also cause the algorithm to waste much computation when a shallow but less forcing proof exists.

Monte Carlo Tree Search \cite{mcts} is a very popular search framework for heuristic game play and single agent search for optimization. The MCTS Solver extension \cite{mcts_solver} adds minimax backpropagation of solved positions in order to solve games. Like PNS, MCTS incrementally builds a tree of explored nodes in memory. However MCTS uses an empirical win rate instead of proof numbers, to guide its move selection according to a formula such as UCT \cite{UCT} or PUCT \cite{puct}. MCTS balances exploitation of moves with the highest win rates against exploration of less visited ones. In each iteration of MCTS, the search tree is expanded with a new leaf node, which is evaluated using either a simulation or a neural network evaluation. Win/loss statistics are propagated along the path back to the root.

$\alpha\beta$ and MCTS can be adapted to find scalar-valued outcomes, while PNS is specialized for binary (win/loss) results.
There are techniques to efficiently specialize $\alpha\beta$ for binary results, and PNS variants for solving scalar-valued games \cite{Saffidine:ECAI2012}. For simplicity, our implementation of EWS solves binary outcome problems. Multi-outcome problems could be solved using multiple searches or extended bookkeeping.

The game of Go is a classic test bed for search algorithm research as it is a popular game with simple rules, yet good play requires complex strategies.
In Go, the two players Black and White take turns placing a stone of their color onto the game board, until both players pass and the game ends. Adjacent stones of the same color are part of the same \textit{block}. A block losing its last adjacent empty point is captured and its stones are removed. Static safety is the process of recognizing without search when points on the board can be guaranteed to count towards one player's score. Such domain-specific knowledge can greatly speed up the search \cite{bss}.

Superko rules in Go prevent infinite loops during games.
Positional superko forbids repeating any previously played board position, and is a commonly used ruleset by Go players. In situational superko, the player to move also has to match in order to repeat a position. Japanese style Go rules address only restricted types of ko, and declare the game as no-result in more complex cases of repetition. Positional or situational superko rules make solving Go more challenging due to the Graph History Interaction (GHI) problem \cite{ghi}. The game history must be considered, which makes it more difficult to reuse previously solved positions. Previous work in solving Go by van der Werf et al. \cite{smallboards} \cite{rectangularboards} used a version of  Japanese rules.  In their MIGOS program, using situational superko instead of Japanese style rules increased the time for solving 4$\times$4 Go from 14.8 seconds to 1.1 hours, more than two orders of magnitude.

The game of Hex is played on a rhombus board with hexagonal cells. As in Go, two players Black and White take turns placing their stones. The goal of the first player (Black) is to connect the top of the board to the bottom using their stones, and the second player (White) tries to connect the left edge to the right edge. Hex has been solved up to the 10$\times$10 board using a Depth-First PNS algorithm (DFPN) \cite{dfpn}, with many Hex specific improvements, in 63 days of parallel computation \cite{10x10hex}. 

\section{Expected Work Search}

\label{sec:ews}

\begin{figure}
    \centering
    \includegraphics[scale=0.7]{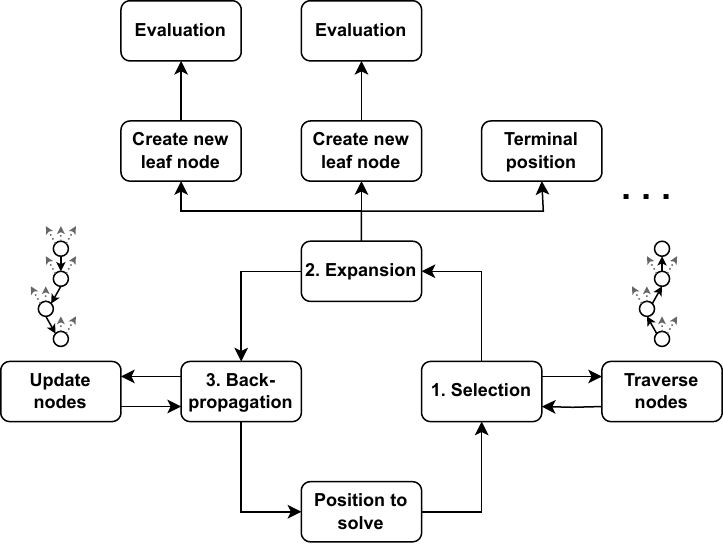}
    \caption{Basic diagram of the structure of Expected Work Search.}
    \label{fig:ews_diagram}
\end{figure}

EWS uses a MCTS-like framework for search. It is designed to minimize a new notion of \textit{Expected Work} (EW), in contrast to algorithms with move selection formulas such as UCT. Figure \ref{fig:ews_diagram} shows the stages of an iteration of EWS: Selection, Expansion, and Backpropagation. EWS uses a negamax formulation with wins and losses defined from the point of view of the player to move.

As in MCTS Solver, repeating these stages grows a search tree of nodes which represent intermediate positions until a complete winning strategy is found for the initial position. The selection stage traverses the search tree to choose a leaf node $X$ for expansion. Expansion adds \textit{all} non-terminal children of $X$ to the search tree and evaluates them (for example by simulation) to initialize their win rate and Expected Work estimates. Backpropagation updates the proofs, child ordering, win rates, and Expected Work of all nodes on the path back to the root. 

EWS requires the following domain-specific functions: getting all legal moves, making and undoing moves, and recognizing terminal positions and their win/loss outcomes. Everything else in the EWS algorithm was designed to be domain agnostic so that it can be easily applied to many different problems.

\subsection{Expected Work}

EWS computes the EW of positions to estimate how much computation they will take to solve.
The Expected Work of a position depends on whether it will be proven to be winning or losing.
For a winning position, only a single winning move needs to be solved, while in a losing position all moves must be solved to be losing.
Therefore, EWS computes separate Expected Work (EW) estimates for both cases for each position.
Let $X$ be a position with $n$ children $C_0,...,C_{n-1}$, arranged in the order in which they will be searched ($C_0$ first, then $C_1$ etc.). Let $W\hspace{-2pt}R(Y)$ be the estimated win rate of $Y$.
Then the Expected Work for  losing and winning  of $X$ is defined recursively as:

\begin{equation}
    EW_{loss}(X) := \sum_{i=0}^n EW_{win}(C_i)
    \label{eq:EWloss}
\end{equation}
\begin{equation}
    EW_{win}(X) := \sum_{i=0}^n ( EW_{loss}(C_i) \cdot \prod_{j=0}^{i-1}W\hspace{-2pt}R(C_j) )
    \label{eq:EWwin}
\end{equation}

The estimated win rate of $X$ is interpreted as the probability that $X$ is a win.
As a base case, a leaf node in the current tree is initialized with a EW heuristic. initialisation options are discussed in Section \ref{sec:ew_init}.

If $X$ is losing, then all its children must be searched and proven to be wins for the opponent.
Equation (\ref{eq:EWloss}) defines $EW_{loss}(X)$ as the sum of $EW_{win}$ of its children.
This corresponds to proof numbers in PNS AND nodes, which are defined as the sum of child proof numbers.

If $X$ is winning, then EWS searches its children in order until the first opponent loss $C_i$ is found, which proves a win for $X$.
Equation (\ref{eq:EWwin}) defines $EW_{win}(X)$ as the sum of all children's $EW_{loss}$, each \textit{weighted by the estimated probability that the search reaches this child} while solving $X$.
For example, suppose $C_0$ has a 80\% win rate and $C_1$ has a 75\% win rate. Then $0.8\cdot0.75=0.6$ is the estimated joint probability that neither $C_0$ nor $C_1$ were solved as losses for the opponent. This is the case when $C_2$ must still be searched, and the EW of $C_2$ is weighted by this probability.
The optimistic approach of PNS, taking the minimum proof number in OR nodes, would correspond to the special case of $WR(C_0)=0$, where the expected work for all other children becomes zero.

The EW computation is based on three simplifying assumptions. First, that win rates correlate with the position actually being winning or losing. EWS does not prescribe how to estimate win rates. A default option using domain-agnostic random simulations is described in Section \ref{sec:win_rate_est}. Second, the definition of Expected Work assumes that the current move ordering among children will remain unchanged. 
However, in practice EWS reorders children continually as the search updates EW.
The third assumption is that the win rates and expected work of siblings are independent. This is a strong assumption, and violated for example in a DAG.
However, our empirical tests in Section \ref{sec:experiments} demonstrate that the above EW definition leads to an efficient search algorithm, validating the usefulness of this framework. 

\subsection{Selection}

\begin{algorithm}
    \setstretch{1.2}
    \caption{SelectBackpropagate $X$: returns whether $X$ is solved and if so whether $X$ is winning}
    \label{alg:select_backprop}
    \begin{algorithmic}[1]
        \State $C$ := $X$.children[0] \Comment{Selection}
        \State Make move $C$.move
        \If{$C$.expanded}
            \State isSolved, isWinning = SelectBackpropagate($C$)
        \Else
            \State isSolved, isWinning = Expand($C$)
        \EndIf
        \State Undo move $C$.move
        \Comment{Backpropagation}
        \If{isSolved and isWinning}
            \State Remove $C$ from $X$.children
            \If{$X$.children is empty}
                \State \textbf{return} true, false \Comment{Solved loss}
            \EndIf
        \ElsIf{isSolved and not isWinning}
            \State \textbf{return} true, true \Comment{Solved win}
        \EndIf
        
        \State Sort $X$.children with (\ref{eq:sort})
        \State Update $X$.winRate
        \State Update $X$.expectedWorkLoss with (\ref{eq:EWloss})
        \State Update $X$.expectedWorkWin with (\ref{eq:EWwin})

        \State \textbf{return} false, false \Comment{Unsolved}
    \end{algorithmic}
\end{algorithm}

The pseudo code for selection can be seen in Algorithm \ref{alg:select_backprop}, lines 1-6. As in MCTS, the selection stage proceeds from the root of the search tree down a path to a leaf. Unlike MCTS, this path simply follows the first-ordered child in the tree according to the current ordering until a leaf node is reached. The child ordering is determined during backpropagation, as described in Section \ref{sec:backpropagation}. Once a leaf node is found, it is expanded as described in Section \ref{sec:expansion}. 

As the in-memory tree is traversed, moves are made accordingly to update the current position. Each node stores a win rate, both expected work values, whether it has been expanded yet, and pointers to its unsolved children.

\subsection{Expansion}
\label{sec:expansion}

\begin{algorithm}
    \setstretch{1.2}
    \caption{Expand $X$: returns whether $X$ is solved and if so whether $X$ is winning}
    \label{alg:expand}
    \begin{algorithmic}[1]
        \State $X$.expanded := true
        \ForAll{legal moves $m$}
            \If{$m$ is a terminal winning move for $X$}
                \State \textbf{return} true, true \Comment{Solved win}
            \ElsIf{$m$ is \textbf{not} a terminal losing move for $X$}
                \State Create node $C$
                \State $C$.expanded = false
                \State $C$.move := $m$
                \State Evaluate $C$.winRate
                \State Evaluate $C$.expectedWorkLoss
                \State Evaluate $C$.expectedWorkWin
                \State Add $C$ to $X$.children
            \EndIf
        \EndFor
        \If{$X$.children is empty}
            \State \textbf{return} true, false \Comment{Solved loss}
        \Else
            \State \textbf{return} false, false \Comment{Unsolved}
        \EndIf
    \end{algorithmic}
\end{algorithm}

When the selection stage reaches a leaf node for position $p$, it is expanded according to Algorithm \ref{alg:expand}.
Expand returns two booleans: whether the node was solved, and if so whether $p$ was winning. All legal moves of the position being expanded are checked to see if they lead to a terminal position. If a move leads to a terminal position which is winning for the player to move in $p$, then the node is solved as a win and the function returns. If a move leads to a losing terminal position, no new node is added. If $p$ has no unsolved children left, it is solved as a loss, otherwise $p$ remains unsolved.

For a move which leads to a non-terminal position, a new leaf node is added as a child of $p$'s node, and its win rate and Expected Work values are initialized by heuristics. This is further described in Sections \ref{sec:win_rate_est} and \ref{sec:ew_init}. 

\subsection{Backpropagation}
\label{sec:backpropagation}

Lines 7-18 in Algorithm \ref{alg:select_backprop} show the pseudo code for EWS backprogagation.  
After the selected child $C$ of node $X$ has finished expanding or backpropagating (lines 4, 6), backpropagation for $X$ proceeds as follows:
If $C$ was solved as a win, it is losing for $X$ and is removed from the (unsolved) children.
If $X$ has no more unsolved children, its result is a solved losing position.
If $C$ was solved as a loss, through backpropagation $X$ becomes a solved win. 

For $X$ that remain unsolved, its unsolved children are re-sorted in ascending order according to the ordering:
\begin{align}
\begin{split}
    A<B&\\
    \Longleftrightarrow \phantom{.}&\\
    EW_{\hspace{-1pt}loss}(A) / (1\hspace{-2pt}-\hspace{-2pt}W\hspace{-2pt}R(A))\hspace{-1pt} < \hspace{-1pt}E&W_{\hspace{-1pt}loss}(B) / (1\hspace{-2pt}-\hspace{-2pt}W\hspace{-2pt}R(B))
\end{split}
\label{eq:sort}
\end{align}
Here, $A$ and $B$ are any two children of $X$. 
We show in the appendix that this child ordering will always minimize the EW of nodes.
Intuitively, children with small EW and low WR are ordered before children with large EW and high WR.
EWS achieves exploration by optimistic initialisation of EW, as described in Section \ref{sec:ew_init}.


The win rate of $X$ is updated after its children are reordered. Details are given in Section \ref{sec:win_rate_est}. Expected work values are updated according to Equations (\ref{eq:EWloss}) and (\ref{eq:EWwin}). Finally, the function return indicates that $X$ is still unsolved.

\section{Implementation}

This section describes choices made for our first full implementation of EWS within the algorithm framework of Section \ref{sec:ews}.

\subsection{Win Rate Estimation}
\label{sec:win_rate_est}

Our implementation of EWS performs win rate estimation with random simulations (a.k.a. rollouts).
A position is simulated by playing random moves until a terminal position is reached, at which point the winner of that simulation is known.
Each node has win and visit counters which are updated during expansion and backpropagation.
Every simulation increments the visit counter of all nodes along the in-tree path chosen in the selection stage, and the win counter for nodes of the simulation winner.
The win rate of a node is simply wins divided by visits.
Wins are initialized to 1, and visits to 2, in order to avoid win rates of exactly 0 or 1. This is required for computing Expected Work (\ref{eq:EWwin}) and the child ordering relation (\ref{eq:sort}).

It is possible to use other methods of win rate estimation with EWS such as a heuristic policy to guide simulations \cite{AlphaGo}, or a machine learned evaluation function \cite{Alphazero} \cite{pcn}.

\subsection{Expected Work initialisation}
\label{sec:ew_init}

EWS does not use an explicit exploration term as in MCTS. It uses optimistic initialisation \cite{optimistic_init} of EW to achieve exploration.
This corresponds to the optimistic initialisation of proof and disproof numbers in PNS \cite{pns}.
In optimistic initialisation, underestimating the EW of new nodes leads to it growing during the initial visits.
In two otherwise similar nodes, the one with fewer visits tends to have a smaller EW estimate and is therefore prioritized in the search.
Typically, a position's optimistic EW initialisation is smaller than the sum of its children's EW.

Our implementation of EWS takes advantage of ``free'' information from the simulations used for win rate estimation to also initialize the Expected Work of new nodes.
EW is initialized as follows:
\begin{equation}
    EW(X) := \sum_{i=0}^{m-1} b(P_i)
\label{eq:EW0}
\end{equation}
where $X$ is the new node being initialized, $P_0,...,P_m$ are the $m$ positions visited during the random simulation, and $b(P_i)$ is the number of legal moves (the branching factor) of $P_i$.
This initialisation method is domain agnostic, gives information about the size of the proof tree required to solve the node, and it reuses simulations used for win rate estimation.
Other plausible initialisation methods for EW include simply initializing EW as 1, and using a learned model to predict proof size \cite{pcn}.

\subsection{Generic Refinements}

We improve the performance of our EWS implementation with the following three game-independent methods:

\begin{enumerate}
    \item Transposition Reduction
    \item Symmetry Reduction
    \item Relevancy Zones
\end{enumerate}


A transposition is an equivalent position that has already been encountered during the search.
We use a transposition table to store the outcomes of solved positions to avoid recomputing such positions.
The transposition table is indexed using hashes that compactly represent positions modulo the table's size. 
We use Zobrist hashing \cite{zobrist} to compute hashes efficiently. We increase the transposition table hit rate using enhanced transposition cutoffs \cite{enhanced_transposition_cutoffs}, which involves performing a 1-ply search for existing transposition entries upon the creation of a new node.

The Graph History Interaction (GHI) problem arises in games such as Go where a position's outcome depends on the game history.
In such domains, this problem occurs when positions that share the same board configuration lead to different game outcomes.
If positions with different histories are treated as identical via a transposition, the search result might become incorrect.
To remedy this issue, we implement the GHI solution proposed by Kishimoto et al.\ \cite{ghi_solved}. Transposition entries are simulated to ensure that the stored outcome can be reached legally within the rules of the game before they are used to replace search.
These simulations incur a memory and computational overhead, but the savings from being able to use a transposition table and re-using a proof rather than needing to rediscover it by search makes up for the overhead.

In many domains, symmetries of positions can be used to improve search efficiency. 
If a position can be mirrored or rotated without affecting the outcome, then solving a position means that all of its symmetrically equivalent positions are also solved.
This allows more positions to be treated as transpositions, increasing the transposition table hit rate and reducing the number of nodes that must be searched.
We also consider symmetries upon creation of new nodes, by omitting any symmetrically equivalent sibling.

EWS uses relevancy zones \cite{rzone} to further reduce the number of nodes required to find proofs. 
Relevancy zones limit the number of moves that must be considered by discarding any moves which cannot be a refutation to a known winning strategy for the opponent. 
Relevancy zones are defined for terminal positions in a domain-specific way. The propagation to earlier positions is domain-agnostic and is handled by EWS in the backpropagation stage. 
See \cite{rzone} for a detailed explanation with concrete examples of relevancy zones, and a proof of correctness.

Two other improvements are used in our implementation: First, if the winner of a starting position is conjectured a priori, then each node's expected win/loss outcome can be set accordingly. 
This allows the search to only compute one EW, by alternating Equations (\ref{eq:EWloss}) and (\ref{eq:EWwin}). This leads to an efficiency gain.
Second, if the search is observed to oscillate rapidly among a small number of children for the same position, its focus can be improved using the $1+\epsilon$ trick \cite{oneplusepsilon}. 

\section{Experiments}
\label{sec:experiments}

\subsection{The 6$\times$6Go Dataset}
\label{go_data}

\begin{table}
    \centering
    \renewcommand{\arraystretch}{1.25}
    \begin{tabular}{|>
    {\centering\arraybackslash}m{50pt}|>{\centering\arraybackslash}m{45pt}|>{\centering\arraybackslash}m{50pt}|>{\centering\arraybackslash}m{45pt}|}
        \hline
        Program & Av. solve time (s) & \# Faster than EWS & \# Timeouts \\
        \hline
        EWS & 2.32 & - & -\\
        \hline
        Go-Solver & 22.63 & 31 (5.2\%) & 11 (1.8\%)\\
        \hline
        EWS-WR & 19.70 & 96 (16.0\%) & 11 (1.8\%)\\
        \hline
        EWS-PS & 19.58 & 171 (28.5\%) & 14 (2.3\%)\\
        \hline
    \end{tabular}
    \caption{Summary statistics of EWS vs. comparison algorithms on 600 tested 6$\times$6 Go positions.}
    \label{tab:go_sum_stats}
\end{table}

To broadly evaluate EWS, we created dataset \textit{6$\times$6Go}
by selecting 600 positions - one from each of 600 6$\times$6 Go games - that are likely to be encountered in a 6$\times$6 Go proof tree.
One player is a strong heuristics-guided agent, representing the good moves made by the winning side.
The opponent plays randomly, representing a fair sample of all legal moves which must be analyzed for the losing side. 
The heuristic player uses six Proof Cost Networks (PCN) \cite{pcn} with different training parameters. PCN are trained to predict minimal proof size, as opposed to the win rate used in AlphaZero \cite{Alphazero}.
We generate 100 games with each PCN to create a dataset with diverse lines of play.
The games in the test set last between 29 and 124 moves, with a median of 40, for a total of 26,885 positions. 
In each game, the earliest position EWS was able to solve in at most 10 seconds was used as a data point in 6$\times$6Go.
All algorithms are tested with a 5 minute limit per position.
Experiments were run on an Intel i7-10510U CPU with 16
GB of RAM.

\subsection{Go Results}

\begin{figure*}
    \centering
    \begin{subfigure}[b]{0.32\textwidth}
        \centering
        \includegraphics[scale=0.5]{"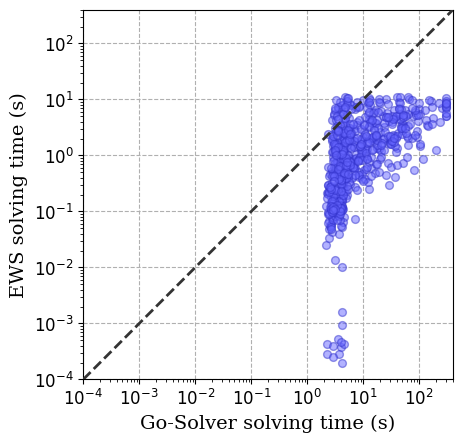"}
        \subcaption{EWS vs. Go-Solver}
    \end{subfigure}
    \begin{subfigure}[b]{0.32\textwidth}
        \centering
        \includegraphics[scale=0.5]{"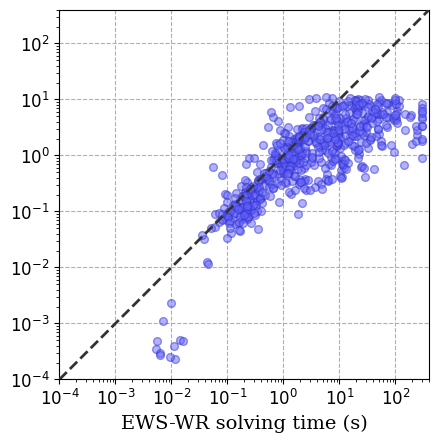"}
        \subcaption{EWS vs. EWS-WR}
    \end{subfigure}
    \begin{subfigure}[b]{0.32\textwidth}
        \centering
        \includegraphics[scale=0.5]{"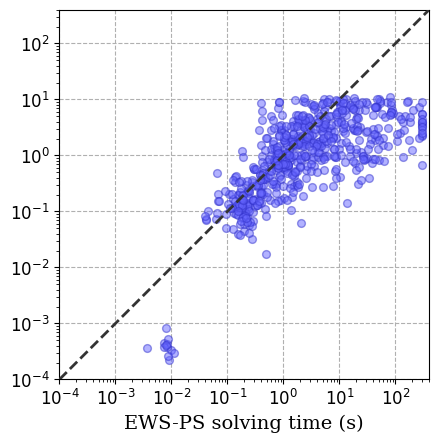"}
        \subcaption{EWS vs. EWS-PS}
    \end{subfigure}
    \caption{ 
    Comparing EWS against Go-Solver, EWS without win rate estimation (EWS-WR), and EWS without proof size estimation (EWS-PS). The y-axis is a log scale of the time EWS took to solve the positions in seconds, and the x-axis is the time it took the other algorithms to solve the same positions. Data points below the diagonal dashed line took less time for EWS to solve than the comparison algorithm.}
    \label{fig:go_scatter_results}
\end{figure*}

\begin{table*}
    \centering
    \renewcommand{\arraystretch}{1.5}
    \begin{tabular}{|>
    {\centering\arraybackslash}m{60pt}||>{\centering\arraybackslash}m{55pt}|>{\centering\arraybackslash}m{55pt}||>{\centering\arraybackslash}m{55pt}|>{\centering\arraybackslash}m{65pt}||>{\centering\arraybackslash}m{55pt}|>{\centering\arraybackslash}m{55pt}|}
        \hline
        Program & 3$\times$3 Time (s) & 3$\times$3 Nodes & 4$\times$4 Time (s) & 4$\times$4 Nodes & 5$\times$5 Time (h) & 5$\times$5 Nodes \\
        \hline
        EWS & 0.053 & 161 & 13.626 & 495,494 & 5.252 & 2,605,781,360\\
        \hline
        EWS-WR & 0.074 & 796 & 40.396 & 1,562,718 & $>$ 24 & -\\
        \hline
        EWS-PS & 0.070 & 232 & 18.320 & 744,169 & $>$ 24 & -\\
        \hline
        Go-Solver & 1.299 & 1,628 & 51.522 & 799,607 & $>$ 24 & -\\
        \hline
        MIGOS SSK & $<$ 3.3 & $\sim$ 25,118 & $\sim$3,960 & $\sim$3,162,277,660 & - & -\\
        \hline
    \end{tabular}
    \caption{The results from solving empty n$\times$n Go boards with EWS, EWS without win rate estimation (EWS-WR), EWS without proof size estimation (EWS-PS) and the $\alpha\beta$ Go-Solver program. 5$\times$5 Time is in hours. MIGOS times and node counts are approximate. All programs use positional superko other than MIGOS which uses situational superko. 3$\times$3 has 8.5 komi, 4$\times$4 has 1.5 komi, 5$\times$5 has 24.5 komi.}
    \label{tab:go_results}
\end{table*}

\begin{table*}
    \centering
    \renewcommand{\arraystretch}{1.5}
    \begin{tabular}{|>
    {\centering\arraybackslash}m{60pt}||>{\centering\arraybackslash}m{55pt}|>{\centering\arraybackslash}m{55pt}||>{\centering\arraybackslash}m{55pt}|>{\centering\arraybackslash}m{55pt}||>{\centering\arraybackslash}m{55pt}|>{\centering\arraybackslash}m{55pt}|}
        \hline
        Program & 4$\times$4 Time (s) & 4$\times$4 Nodes & 5$\times$5 Time (s) & 5$\times$5 Nodes & 6$\times$6 Time (h) & 6$\times$6 Nodes \\
        \hline
        EWS & 0.002 & 283 & 0.253 & 37,034 & 0.422 & 93,963,192\\
        \hline
        Enhanced AB & 0.004 & 1,673 & 3.935 & 698,402 & $>$ 24 & -\\
        \hline
        Morat MCTS & 0.035 & 4,644 & 29.669 & 3,554,546 & $>$ 24 & -\\
        \hline
        Morat PNS & 0.054 & 8,871 & 758.102 & 154,539,591 & $>$ 24 & -\\
        \hline
    \end{tabular}
    \caption{The results from solving the empty n$\times$n Hex board without extensive domain-specific knowledge with EWS, an enhanced $\alpha\beta$ program, Morat MCTS, and Morat PNS. 6$\times$6 Time is in hours.}
    \label{tab:hex_results}
\end{table*}

\begin{table*}
    \centering
    \renewcommand{\arraystretch}{1.5}
    \begin{tabular}{|>
    {\centering\arraybackslash}m{60pt}||>{\centering\arraybackslash}m{55pt}|>{\centering\arraybackslash}m{55pt}||>{\centering\arraybackslash}m{55pt}|>{\centering\arraybackslash}m{55pt}||>{\centering\arraybackslash}m{55pt}|>{\centering\arraybackslash}m{55pt}|}
        \hline
        Program & 6$\times$6 Time (s) & 6$\times$6 Nodes & 7$\times$7 Time (s) & 7$\times$7 Nodes & 8$\times$8 Time (s) & 8$\times$8 Nodes \\
        \hline
        EWS with knowledge & 0.005 & 26 & 0.588 & 2,318 & 234.449 & 555,158\\
        \hline
    \end{tabular}
    \caption{The results from solving the empty n$\times$n Hex board using domain-specific knowledge with EWS.}
    \label{tab:knowledge_hex_results}
\end{table*}

To evaluate EWS, we compare it against three baselines: Go-Solver, EWS-WR (EWS without win rate estimation), and EWS-PS (EWS without proof size estimation). Go-Solver is an iterative deepening $\alpha\beta$ algorithm built on the open source Fuego framework \cite{fuego}.
Go-Solver is the strongest existing Go solving program using positional superko rules we are aware of, apart from EWS.
For these experiments, Go-Solver uses the exact same static safety Go knowledge implementation as EWS, so it statically recognizes the same wins and losses.

In EWS-WR we replace Equation (\ref{eq:EWwin}) with $EW_{win}:=\min(EW_{loss}(C_i))$ to take the minimum child EW instead of a weighted average, and set all node's WR to 0.
This results in a negamax PNS algorithm which prioritizes positions closest to being solved.

In EWS-PS we order children of each node according to the UCT formula \cite{UCT}. 
This algorithm is similar to MCTS solver \cite{mcts_solver}, with minor implementation differences.

Figure \ref{fig:go_scatter_results} compares EWS against the three baselines on the 6$\times$6Go dataset.
Each scatter plot uses doubly-logarithmic scales, with EWS solving time on the y-axis, and the other algorithm's solving time on the x-axis. 
For data points below the diagonal, EWS is faster.

Table \ref{tab:go_sum_stats} shows the summary statistics of these experiments. 
Go-Solver has a one second initialisation overhead, and was generally slower for harder positions.
Go-Solver's average solving time is 9.5$\times$ larger than the average time it took EWS to solve the same positions.

The PNS-like EWS-WR is a respectable solving algorithm even without win rate estimation, with performance comparable to Go-Solver. 
On average EWS-WR took 8.5$\times$ longer than EWS to solve the same positions.
The MCTS-like EWS-PS had the fastest average solving time of the three competitors, and solved the most positions faster than EWS.
However, EWS-PS also exceeded the 5 minute time out period on the most positions.
This experiment shows that 
the performance of EWS degrades when either one of its two main attributes is removed from the search.

\subsubsection{Solving Small Square Go Boards}

We compared EWS, EWS-WR, EWS-PS, and Go-Solver against the state of the art by solving the empty square Go boards of size $3\times3$ up to $5\times5$. 
State of the art is represented by the MIGOS program, which is accepted as the strongest previously published Go solver \cite{smallboards} \cite{rectangularboards}. 
Results are shown in Table \ref{tab:go_results}. 

While the strongest results for MIGOS are for Japanese style rules, it was also tested using situational superko (SSK) rules up to the empty 4$\times$4 board, shown in Table \ref{tab:go_results}. Our Go implementation uses positional superko rules, which differ slightly from SSK as they forbid any positional repetition regardless of the player to move. Both rulesets lead to GHI problems \cite{ghi}, making them comparable for solving.

In our experiments EWS outperforms the other tested algorithms and was the only program to solve 5$\times$5 Go, with 24.5 komi, within 24 hours. For the first time, 5$\times$5 Go has been solved with the positional superko ruleset, confirming the full-board first player win for Japanese rules. We naively parallelized the programs when solving 5$\times$5 Go by running 6 parallel processes to solve the positions resulting from the center opening move for Black and the 6 symmetrically unique responses from White.
No other experiments reported here used any parallel computing.

\subsection{Hex}

To show the generality of EWS, we also evaluated it on the game of Hex. Table \ref{tab:hex_results} shows the results of solving empty n$\times$n Hex boards using little Hex-specific knowledge. We compare EWS against the publicly available Morat program \cite{morat} which includes PNS and MCTS algorithms to play and solve Hex. Morat PNS implements the Depth First Proof Number search variant of PNS, along with refinements such as the $1+\epsilon$ trick \cite{oneplusepsilon}. We also implemented an enhanced $\alpha\beta$ algorithm to solve Hex. All programs tested in the results shown in Table \ref{tab:hex_results} have the same limited Hex specific knowledge.

The results in Table \ref{tab:hex_results} show that EWS is faster and creates smaller proof trees than any of the other tested algorithms. For solving 5$\times$5 Hex, EWS was 15.5$\times$ faster than enhanced $\alpha\beta$, 117.3$\times$ faster than Morat MCTS, and 2996.5$\times$ faster than Morat PNS. EWS was the only program to solve 6$\times$6 Hex in the 24 hour time limit allotted. 

6$\times$6 Hex is a smaller problem than the state of the art, which has solved 10$\times$10 Hex \cite{10x10hex}.
To show the importance of domain-specific knowledge in Hex, we implemented Hex specific knowledge for EWS as shown in Table \ref{tab:knowledge_hex_results}: 
basic virtual connections which identify terminal positions earlier, and fill-in / inferior cell patterns which prune dominated moves \cite{7x7hex}.
Compared to \cite{10x10hex}, our current Hex knowledge implementation lacks the more complicated virtual connections and fill-in / inferior cell patterns.

Our results emphasize the importance of domain-specific knowledge for solving large Hex problems. In Table \ref{tab:hex_results}, EWS explored over 93 million nodes to solve 6$\times$6 Hex with no domain knowledge. 
With basic Hex-specific knowledge, EWS takes only 26 nodes, and
 solves 8$\times$8 Hex in less than 4 minutes\footnote{Currently EWS is restricted to boards with at most 64 cells due to code base limitations.}. 

\section{Related Work}

The PN-MCTS algorithm \cite{mcts_pns} is a game playing program which has shown favourable results compared to unenhanced MCTS on a number of games.
PN-MCTS is MCTS with a modified UCT formula that uses proof and disproof numbers to help guide move selection. 
Proof and disproof numbers are used to rank children, which is used in a normalized term added to the UCT formula which influences move selection.
All information in the proof and disproof numbers other than the relative ordering of children's proof numbers is discarded.
A major difference from EWS is that PN-MCTS keeps its win rate and proof number-based heuristics separate, and only combines them through adding terms to UCT during move selection.
In contrast, EW integrates win rate and proof number-like information into a single expected work estimate, which is computed recursively.
EWS is also able to consider the estimated proof sizes of positions rather than just considering children's relative ordering.
Another main difference between PN-MCTS and EWS is that PN-MCTS was designed to be a game-playing program, while EWS is a game solver.
Since the published PN-MCTS algorithm is not enhanced to solve games, it makes fair empirical comparisons difficult, and we leave this direction to future work.

Product propagation \cite{pp} modifies PNS by backing up heuristic estimates of the likelihood of winning rather than proof and disproof numbers. This modification replaces the normal proof size estimation which occurs in PNS, with the goal of achieving a stronger heuristic for the search. Product propagation search was empirically shown to outperform $\alpha\beta$ search and PNS in some of the games tested.

Previous work by van der Werf et al.\ on solving Go has solved rectangular boards up to 5$\times$6 and 4$\times$7 \cite{smallboards} \cite{rectangularboards} using Japanese style rules. Their MIGOS solver uses iterative deepening $\alpha\beta$ with handcrafted heuristics and Go-specific knowledge. Japanese style rules were used in order to avoid the Graph History Interaction problem \cite{ghi}. With situational superko, MIGOS was able to solve up to the empty 4$\times$4 board.

The first computer solution of 7$\times$7 Hex by Hayward et al.\ relied on advances in the Hex-specific knowledge of virtual connections and move domination \cite{7x7hex}. Improvements on inferior cell analysis and pattern decomposition allowed 8$\times$8 Hex to be solved \cite{8x8hex}. Increasing the strength of virtual connections with captured cell knowledge and switching to a DFPN algorithm allowed solving 9$\times$9 Hex \cite{9x9hex}. Finally, a large-scale parallel computation solved 10$\times$10 Hex \cite{10x10hex}.

\section{Conclusions and Future Work}

EWS successfully combines win rate and proof size estimation in an efficient search algorithm. Our experiments show that EWS can outperform $\alpha\beta$, PNS, and MCTS based programs in Go and Hex. We present the new result of solving 5$\times$5 Go with positional superko, and demonstrate that EWS can solve 8$\times$8 Hex when given basic Hex-specific knowledge.

For future work, we plan to test machined learned evaluation functions for win rate estimation \cite{AlphaGo} \cite{Alphazero} and EW initialisation \cite{pcn}. We will continue to develop our Hex specific knowledge with the goal of obtaining new results in solving Hex. We also plan to evaluate EWS on domains other than Go and Hex. Finally, we hope to investigate translating EW into a concrete prediction of the resources it will take to solve problems.

\bibliographystyle{acm}
\bibliography{mybib.bib}

\appendix
\section{Proof of Optimality for Node Move Ordering}
\label{sec:appendix_proof}
Proof sketch:
Claim: Nodes $n$ should be sorted as $(n_1,...,n_k)$ in increasing order of $f(n) = EW_{loss}(n) / (1-W\hspace{-2pt}R(n))$.
We assume $EW_{loss}(n) \geq 0$ and $0 < W\hspace{-2pt}R(n) < 1$ for all nodes $n$. 
Nodes have a wins counter initialized to 1, and a visits counter initialized to 2, and since the win rate of a node is wins / visits, the win rate of a node will never equal zero or one.

Proof by contradiction, assume that a different ordering has strictly smaller total EW. Then two nodes $A$, $B$ must exist s.t. $A = n_i, B = n_{i+1}$, but $f(A) > f(B)$.
We show that changing the order of $A$ and $B$, such that  $B = n_i, A = n_{i+1}$, leads to a strictly smaller EW.

$EW_{win} = \sum_{i=0}^n (EW_{loss}(n_i) \cdot \prod_{j=0}^{i-1}W\hspace{-2pt}R(n_j))$

If we swap $A$ and $B$, all but the two terms with $EW_{loss}(A)$ and $EW_{loss}(B)$ stay the same.
With ordering $A$ before $B$, and setting $p = \prod_{j=0}^{i-1}W\hspace{-2pt}R(n_j) > 0$, $a = EW_{loss}(A)$, $b = EW_{loss}(B)$, these two terms are:
$$EW_{A,B} = ap + bp \cdot W\hspace{-2pt}R(A)$$
With ordering $B$ before $A$, these two terms are:
$$EW_{B,A} = bp + ap \cdot W\hspace{-2pt}R(B)$$
We show that the difference $EW_{A,B} - EW_{B,A} > 0$:
$$EW_{A,B} - EW_{B,A} = $$
$$ap + bp \cdot W\hspace{-2pt}R(A) - (bp + ap \cdot W\hspace{-2pt}R(B)) =$$
$$ap - ap \cdot W\hspace{-2pt}R(B) - bp + bp \cdot W\hspace{-2pt}R(A) =$$
$$ap \cdot (1 - W\hspace{-2pt}R(B)) - bp \cdot (1 - W\hspace{-2pt}R(A)) =$$
$$p \cdot (1-W\hspace{-2pt}R(A)) \cdot (1-W\hspace{-2pt}R(B)) \cdot (a / (1 - W\hspace{-2pt}R(A)) - b / (1 - W\hspace{-2pt}R(B)) =$$
$$p \cdot (1-W\hspace{-2pt}R(A))\cdot (1-W\hspace{-2pt}R(B))\cdot (f(A) - f(B)) > 0$$
Since all terms are positive.

\end{document}